\documentclass{article}
\usepackage{arxiv}

\usepackage[utf8]{inputenc} 
\usepackage[T1]{fontenc}    
\usepackage{hyperref}       
\usepackage{url}            
\usepackage{booktabs}       
\usepackage{amsfonts}       
\usepackage{nicefrac}       
\usepackage{microtype}      
\usepackage{xcolor}

\title{Why Talking about ethics is not enough: a proposal for Fintech's AI ethics}

\author{
  Cristina Godoy Bernardo de Oliveira\thanks{On a sabbatical leave sponsored by the Institute of Advanced Studies of the University of São Paulo.} \\
  Faculdade de Direito de Ribeirão Preto\\
  University of São Paulo\\
  Ribeirão Preto, SP, BRAZIL \\
  \texttt{cristinagodoy@usp.br } \\
   \And
 Evandro Eduardo Seron Ruiz \\
  Department of Computing and Mathematics\\
  FFCLRP -- University of São Paulo\\
  Ribeirão Preto, SP, BRAZIL \\
  \texttt{evandro@usp.br} \\
}

\begin{document}
\maketitle

\begin{abstract} 
As the potential applications of Artificial Intelligence (AI) in the financial sector increases, ethical issues become gradually latent. The distrust of individuals, social groups and governments about the risks arising from Fintech's activities is growing. Due to this scenario, the preparation of recommendations and Ethics Guidelines is increasing and the risks of being chosen the principles and ethical values most appropriate to companies is high. Thus, this exploratory research aims to analyze the benefits of the application of the stakeholder theory and the idea of Social License to build an environment of trust and for the realization of ethical principles by Fintech. The formation of a Fintech association for the creation of a Social License will allow early-stage Fintech to participate from the beginning of its activities in the elaboration of a dynamic ethical code and with the participation of stakeholders.
\end{abstract}


\section{Introduction}

As the potential applications of Artificial Intelligence (AI) in the financial sector increases, ethical issues become gradually latent. The distrust of individuals, social groups, and governments about the risks arising from Fintech's activities is growing. Due to this scenario, the preparation of recommendations and Ethics Guidelines is increasing, and the risks of being chosen the principles and ethical values most appropriate to companies is high. Thus, this exploratory research aims to analyze the benefits of applying the stakeholder theory and the idea of Social License to build an environment of trust and for the realization of ethical principles by Fintech. The formation of a Fintech association for the creation of a Social License will allow early-stage Fintech to participate from the beginning of its activities in elaborating a dynamic ethical code and with the participation of stakeholders.

The financial sector is one of the fields that most seeks technological solutions and applications of Artificial Intelligence in its services. In this scenario, Fintech (Financial technology) quickly presents solutions in more than 19 business areas of the financial sector. Although many Fintech swells, few could climb or remain in this competitive market, mainly due to legal regulations that are costly to early-stage Fintech. In addition, Fintech are new companies that do not yet have a strong brand, so consumer distrust is high. Thus, it is necessary to think about solutions so that Fintech could survive and create answers that help the financial sector and society. 

Hence, our article suggests that the creation of an association formed by Fintech from several countries allows stakeholders' participation for the elaboration, monitoring, and correction of ethical principles followed by Fintech is a way to create an environment of trust between companies and consumers. Also, a Social License is a way to increase Fintech's credibility and strengthen these companies' dialogue with the government to solve legal problems more quickly. 

Finally, the word ethics must have meaning and be applied within the data-driven industry. Artificial Intelligence (AI) presents several solutions and facilities to human problems, and it is necessary to increase the dialogue between society and technology companies to create a trustworthy environment. 

\section{Artificial Intelligence}
Creating and applying non-biological intelligent systems has been a long pursued task as summoned by Lee Spector~\cite{spector2006evolution}. Spector also recalls that artificial intelligence (AI) was not really founded at the famous Dartmouth College conference~\cite{crevier1993ai}. The rumors of artificial beings endowed with intelligence since the 19th century when William Paley argued ``that intelligent designers are necessary for producing complex adaptive systems''~\cite{shapiroPaley}. Spector also remembers that Charles Darwin refuted this idea around 1859, showing that complex and adaptive systems, nowadays called intelligent systems, can arise naturally from a process of selection.

Nowadays, abstract models of cognition under artificial agents and the software instantiations of such models are what has been called `Cognitive Architectures'~\cite{LIETO20181}. These cognitive architectures aim to produce artificial systems that exhibit intelligent behavior analog to human in every detail, including the ability to take unsupervised decisions ensuring the safety of their behavior, and more than that, they should also be explicitly ethical~\cite{VANDERELST201856}.

The current evolution of AI systems has brought in an extensive discourse on AI ethics. Hagendorff~\cite{hagendorff2020ethics} brings an updated analysis and comparison of 22 guidelines, namely normative principles and recommendations aimed to deprive AI systems of any disturbing new AI technologies. Earlier, Jobim and co-authors~\cite{jobin2019global} analyzed whether a global agreement about the major ethical questions was emerging based on some AI guidelines. Their study revealed a global convergence around five ethical principles, which are: 1) Transparency; 2) Justice and fairness; 3) Non-maleficence; 4) Responsibility, and; 5) Privacy. These principles will be discussed directly or indirectly in this paper, but before that, we shall recall the main tasks we can find AI methods being applied. Together, principles and applications might bring a clearer picture of the eventual distress employed by AI techniques.   

What we humans may think about AI, the intelligence exhibited by machines, is not the same as the natural intelligence demonstrated by animals and humans as it involves emotions and consciousness. The AI of today is manifested by what we call `Machine Learning' (ML) methods. Machine Learning is a technology where computers can learn from examples of data associations. An ML task results from an evaluation, sometimes a prediction or inference, of many sampled data. Machine learning tasks rely on learning from patterns in data and not by explicitly programming a computer. This initial amount of data presented to the ML algorithms is usually called `training data.'  It is used to build a working model. This model drives the prediction or the inference engine. Building a model from training data is what we call `supervised learning', and although primarily used by ML approaches, it is not the only one.

Applications for ML methods vary as much as human creativity might take it, but the majority of the ML methods can be resumed as the following tasks: 

\begin{itemize}
    \item Binary classification, which is the task of classifying the elements of a set into two distinct groups. Quality control is a well known binary classification test;
    \item Multiclass classification is the task of classifying the elements into three or more classes. Language identification systems are a suitable example of multiclass classification
    \item Regression is a statistical technique for estimating the relationships among variables. Predictions from historic data most often are based on regression algorithms; 
    \item Clustering is a technique used to group multi-dimensional data into closely related groups. The task of finding similar users on social networks based on the contents of their posts is an example of clustering;
    \item Ranking may also be a classification task as it involves scaling and rating systems. It is a task generally applied to information retrieval, e.g., ranking the web links retrieved by a search algorithm looking at web pages. A recommendation may also be thought of as a ranking system, and finally;
    \item Forecasting, as a process for making predictions of future data based on previous or present samples.
\end{itemize}

This previous compilation of ML tasks is here to invite the reader to foresee that some ethical issues in artificial intelligence might be the work of one or a group of these algorithms. It might be now clear that the expected predictability of these methods as they rely on their `knowledge' and trained data is positively related to the loss of predictable physical work jobs such as soldering and packing in an assembly line. Racism and AI bias can be understood as an improper balance of the data samples. Bias is a part of life. Bias origin accompanies a diversity of data that each group is represented by. This same data is the hallmark that naturally divides them by the algorithms.

Finally, the existing discrimination in society could not be eliminated through corrections of algorithms. The data used to reflect an image of a particular society's past, and the change depends on state intervention and public policies. On the other hand,  the construction of a trustworthy AI depends on companies' initiative to avoid the perpetuation of discrimination. In this way, our research will focus on the case of financial technology (FinTech) that use AI for prusesling and granting credit, directly impacting people's lives. Furthermore, we will propose a model of association between Fintech and the issuing of a certification so that companies' performance for the implementation of ethical principles is carried out in a more efficient way.

\subsection{AI and Fintech}

AIDS is an useful concept for this article brought by Longbing Cao~\cite{cao2020ai} that means:
\begin{quote}
AIDS broadly refers to both classic AI areas including logic, planning, knowledge representation, modeling, autonomous systems, multiagent systems, expert system (ES), decision support system (DSS), simulation, pattern recognition, image processing, and natural language processing (NLP) and modern AI and DS areas such as representation learning, machine learning, optimization, statistical modeling, mathematical modeling, data analytics, knowledge discovery, complexity science, computational intelligence, event and behavior analysis, social media/network analysis, and more recently deep learning and cognitive computing. In contrast, we broadly refer finance to areas including capital market, trading, banking, insurance, leading/loan, investment, wealth management,
risk management, marketing, compliance and regulation, payment, contract, auditing, accounting, financial infrastructure, blockchain, financial operations, financial services, financial security, and financial ethics.
\end{quote}

In order to comprehend the expression `Smart Fintech', Longbing Cao~\cite{cao2020ai} stated that Fintech is a keystone of ``synthesizing, innovating and transforming financial services, economy, technology, media, communication, and society broadly driven by AIDS techniques''. Besides, AIDS techniques represent the combination of the classic AI areas, such as logic, planning, autonomous systems, expert system (ES), support system (DSS,) natural language processing (NLP), among others, and also, modern AI and DS areas (representing learning, machine learning, optimization, data analytics, etc.). Therefore, EcoFin (economics and finance) and the new generation of FinTech are based on AIDS techniques, and the result of this is called \textbf{smart FinTech}.

Longbing Cao also identified 19 different Fintech business areas, and our research focuses on economic-financial services, such as banking, insurance, lending, financing, and crowdfunding services. Within this business area, there are several challenges to be faced by smart FinTech. Our article will seek solutions concerning ``human and social complexities: such as AIDS techniques for modeling and managing the diversity and inconsistency of participant’s cognitive, emotional and technical capabilities and performance, and for enabling effective communications, cooperation and collaboration within a department and between stakeholders(\ldots)''~\cite{cao2020ai}.

FinTech represents a relevant strategy for banks and start-ups to adapt to the changes that technology is generating in society~\cite{belanche2019artificial} (p.~1411). Thus, it is not restricted to e-banking and consumer digitalization, but the strengthening of relations between users and banking services. Thus, AI represents an opportunity for modifying the financial industry and increasing revenue~\cite{park2016robo}.

For Fintech, data has the following functions: providing data to its customers and providing solutions to the Financial Service Institutions (FSIs) to improve their services and to create new products and services~\cite{gai2018survey}. Consequently, rapid data analysis and proper information management are crucial for developing new services for FSIs. Therefore, data analysis and artificial intelligence are fundamental in this scenario.

The financial sector has always sought innovations to improve its services and create new opportunities, but, currently, there is a focus on the use of new technologies to provide solutions and the creation of new financial services. Thus, Fintech uses big data analytics, artificial intelligence (robot advisors, credit score, credit risk, etc.), and blockchain technology to meet the high demand for new products, solutions, and services~\cite{giudici2018fintech}.

Fintech comes up with new services and solutions for FSIs in an environment of excessive legal regulation, causing the emergence of `Regtech' (Regulatory Technology) firms. These Regtech provide legal solutions that can help banks and other intermediaries so that they do not violate the legislation and could manage risks more effectively~\cite{bofondi2017thebig}. Also, Fintech faces two other challenges: security problems and lack of confidence in the data used. Thus, building a relationship of trust between users and Fintech is essential for activities to be expanded and new services to be offered~\cite{mention2019future} (p.~61).

Fintech are suitable for our analysis proposed in this article because they have some difficulties that can be overcome by employing a Social License, such as: they have difficulty scaling the business due to lack of trust from its users, lack of brand awareness,  ``an established distribution infrastructure, capital, and regulatory compliance expertise that, historically, are the strengths of incumbent firms''~\cite{bose2018world} (p.~10).

Besides, Fintech are based on the experience of consumers of their services, so they create an environment where there is trust through ``principles of personalization, quick response (speed), relevance, transparency, and
seamless delivery''~\cite{bose2018world} (p.~14). Thus, conducting tests simulating in real-time and in a realistic manner its operations is essential for the development of its platform, and it is only possible through collaborations with other companies in a favorable regulatory environment~\cite{mention2019future}. As a result, testing with live simulations increases consumer confidence, and automated discriminatory decisions could be corrected, for example.

On the other hand, it is difficult to identify the limits of this collaborative relationship with other companies, as the early-stage Fintech is also a competitor~\cite{zetzsche2017regulating} (p.~07).  As 95\% of Fintech fail to scale the business due to the absence of strategic collaboration~\cite{mention2019future}, it is noted that this is the favorable scenario for the development of a Fintech association aimed at conducting ML tests, consumer awareness regarding the use of your data by Fintech, participation of all stakeholders in the development of ethical principles and values, etc. Thus, collaboration is necessary for Fintech's survival, and our proposal could contribute to building scalable business.

In the end, tests related to Fintech are changing because regulatory sandbox initiatives have emerged in several countries, allowing the development of a safe environment so that early-stage Fintech could carry out their tests without the need to obtain a full license that could impede due to legal costs required to obtain it~\cite{dostov2017regulatory}. Consequently, regulatory sandboxes help develop a system of ``startup-friendly, cross-sectoral and global ecosystem''~\cite{mention2019future}. Thus,  regulatory sandbox and data analytics will provide a trustworthy legal and economic environment through data processing transparency.

\subsection{Apple Card Case and Gender Discrimination}

In the United States, there is a US Equal Credit Opportunity Act (ECOA) that prohibits collecting and using gender data for credit rating assessment. As a result, Fintech does not use gender data for ML training in the US, which causes discriminatory decisions like Kelley and Ovchinnikov~\cite{kelley2020anti} seek to demonstrate.

Furthermore, David Heinemeier Hansson, a software developer, demonstrated that Apple Card~\cite{vigdor2019apple} was a `sexist program' because his credit line was 20 times higher than his wife. After this case, the debate about a discriminatory algorithm increased, and research was developed to face gender discrimination in ML training.

Consequently, Kelley and Ovchinnikov~\cite{kelley2020anti} studied the reduction of discrimination based on gender for the granting of credit by Fintech. Thus, three types of anti-discrimination legislation were analyzed:

\begin{description}
\item[Level 1] Legislation that authorizes data collection and the use of gender for training machine learning (ML) (Example: Singapore);
\item[Level 2] Legislation that authorizes data collection regarding gender, but does not authorize use for ML training (Example: European Union (EU));
\item[Level 3] Legislation that does not authorize the collection and the use of gender data for ML training (Example: United States (US)).
\end{description}

Comparing statistical and other ML models, Kelley and Ovchinnikov~\cite{kelley2020anti} (p.~06) stated that the withdrawal of gender data interferes in various stages of the ML modeling process, which are not captured by the omitted variable bias (OVB) such as feature engineering (the use of gender data creates new features), algorithm selection (outperformance of a random forest model, e.g.), feature selection (e.g., the use of SHAP values and SHAP interaction values demonstrated that ``gender-reliant features are on average 19 times more important for women compared to men''~\cite{kelley2020anti} (p.~07), and also hyperparameter selection. They concluded that the gender data and engineered features impact the set of hyperparameters, which impacts the model predictions.

Consequently, Kelley and Ovchinnikov~\cite{kelley2020anti} carried out some studies to analyze how it is possible to reduce discrimination by gender according to existing legislation and presented four approaches:

\begin{enumerate}
    \item Down-sampling the training data to rebalance gender: is possible in the Level 1 and 2 of anti-discriminatory legislations and  ``results in -9.43\% discrimination, - 174 bps predictive quality, -0.06\% avg. profitability''~\cite{kelley2020anti} (p.~08);
    \item Gender-aware hyperparameter turning: is possible in the Level 1 and 2 of legal regulation and the results are ``-24.56\%, -280 bps predictive quality, -0.11\% average profitability''~\cite{kelley2020anti} (p.~07);
    \item Up-sampling the training data to rebalance gender: is feasible in the Level 1 and 2 of anti-discriminatory laws and results are ``-2.55\% discrimination, no significant change to predictive quality and - 0.04\% average profitability''~\cite{kelley2020anti} (p.~08);
    \item Probabilistic gender proxy modeling: is possible to be applied it in Level 3 but the US legislation  prohibits. The results are ``-62.08\% discrimination, +0.01\% average profitability and no significant change to predictive quality''~\cite{kelley2020anti} (p.~08).
\end{enumerate}

The research by Kelley and Ovchinnikov~\cite{kelley2020anti} demonstrates that legislation can impact discrimination for the granting of credit to Fintech since excluding gender data does not correct discrimination by gender but intensifies it. 
In this way, Aristotle's statement~\cite{aristotle1983ethics} (p.~160) is extremely current to face the challenges arising from the use of artificial intelligence in the financial field:

\begin{quote}
Thus injustice, as we say, is both an excess and a deficiency, in that it chooses both an excess and a deficiency--in one's own affairs choosing excess of what is, as a general rule, advantageous, and deficiency of what is disadvantageous; in the affairs of others making a similarly disproportionate asymmetry, though in which way the proportion is violated will depend upon circumstances\ldots
\end{quote}

Hence, we observed that legislation could impact ML applications' results, and it is necessary to update it when the occurrence of discrimination is detected due to legal restrictions. Suppose Fintech organizes itself in an association to apply the ethical principles and values of society. In that case, it is possible to maintain constant contact with legislators to avoid what occurred in the Apple Card case, for example. Thus, we will present below the Social License for Fintech to help realize ethical decisions by Fintech.

\section{Discussion}
Are only the AI systems that may have ethical issues? Have we assumed that previously to AI, all computer systems and algorithms were fair and ethical? Let us take a simple but very effective ML approach called Naïve Bayes, here called NB. NB is a family of `probabilistic classifiers' based on the Bayes' theorem. The 18th-century British mathematician Thomas Bayes created a formula for determining a conditional probability used to measure the likelihood of an outcome occurring based on previous events. What is `intelligence' there for just calculating probabilities? How difficult is it to count events? How could an 18th-century formula be avoided from use during the last 30 or 40 years? Person profiling might be an application of the Bayes theorem. For instance, knowing that someone is a quiet person and prefers to spend his/her free time reading the newspaper, one might guess whether he/she is a librarian or a salesperson. Computer codes are pervasive. They are everywhere. AI systems can be programmed from various methods, from sophisticated ones to the more simple \texttt{if-them-else} commands; however, unlikely this latter approach might be. Are the consumers trained to recognize ethical flaws in these systems? Are the programmers trained for challenges like this one?

\subsection{Codes of digital ethics: risks and challenges }

The term `ethics' could lose its meaning if we do not connect it with action. Ethics guides us to decide better in a context based on available information. Thus, we need to analyze ethical values and principles thinking about how to act to apply them. Therefore, we will talk about the risks of talking and writing about ethics when we ignore practicing ethical values and principles. 

Between 2017 and 2018, more than 70 recommendations on ethics of AI were published~\cite{winfield2019updated}. Luciano Floridi~\cite{floridi2019translating} highlighted five unethical risks due to the increase of recommendations related to ethics in artificial intelligence applications: 
\begin{enumerate}
  \item Ethics shopping: it is a practice used to justify itself concerning the decisions, choices, and processes adopted by a company. In this way, the most appropriate principles for your objectives are chosen as a `menu' of principles and values. In this way, instead of companies looking for behavioral changes, alterations, and corrections of algorithms and changes in data processing, they aim at the selection of principles that legitimize their conduct;

  \item Ethics blue washing: it is about adopting cosmetic measures to appear the existence of ethical concerns by technology companies. Thus, there is a focus on using advertising and marketing to cover the company in the appearance of being concerned with ethical issues. However, in the field of action, nothing is done, as the implementation of effective measures to correct processes or decisions can generate the expense of many resources;

  \item Ethics lobbing: refers to the exploration of self-regulation through codes of ethics to delay or prevent the drafting of relevant legislation, to justify a limitation of compliance, and to weaken enforcement. Companies do not see what the benefits of acting ethically regarding their lobbying policy~\cite{benkler2019don} and prefer to ``avoid good and necessary legislation (or its enforcement) about design, development, and deployment of digital processes, products, services, or other solutions''~\cite{floridi2019translating};

  \item Ethics dumping: refers to the export of unethical research activities to countries where they do not have strong legislation that prohibits the conduct of these practices and the importation of the results from this research to the country of origin of the company that conducted it. The term `ethics dumping' was introduced by the Science with and for Society Unity of the European Commission in 2013. Finally, there are two reasons for the export of unethical practices: intentional exploitation from high-income to low or middle-income countries and exploitation due to lack of knowledge and ethical concerns~\cite{schroeder2019ethics} (p.~02); and 

  \item Ethics shirking: it is about carrying out few activities aimed at the application of ethical principles and values when the return is low, and the cost of responsibility is transferred to third parties. The term shirking comes from the financial market and means ``the tendency to do less work when the return is smaller''~\cite{nasdaq2021shirking}.
\end{enumerate}

Due to these five risks arising from the excess of codes of ethics, Floridi~\cite{floridi2019translating} states that there are some strategies to face these malpractices:
\begin{itemize}
  \item For ethics shopping concerns, a clear ethical guide could be an effective initiative (e.g., Ethics Guidelines for Trustworthy AI  -- EU)~\cite{ethicsGuidelines}, widely disseminated and accepted by many companies so that malpractice is easily detected;

  \item For ethics blue washing and dumping, a good strategy could be strengthening transparency and education because this initiative could help identify companies that do not adopt ethical practices. Besides, a certification system for digital products and services can help implement this strategy;

  \item For ethics lobbying, a great strategy is to enact clear legislation and make it effective. Furthermore, the exposition of companies that practice lobbying to delay the approval of laws and the differentiation between these companies from those that have efficient self-regulation is essential;

  \item For ethics shirking, the strategy consists of finding the malpractice source and reallocating the distribution of responsibilities.
\end{itemize}

Although strategies to address these malpractices are pointed out, there is a lack of a model capable of facing digital companies' ethical problems. Thus, we will present an association model with its roots in social license and open software initiatives to implement strategies aimed at combating unethical conduct.

\subsection{Social license for data-driven industry }

Accounting provides a relevant mechanism for social control~\cite{brown2006approaches} (p.~107). The consumers have the `right to know'~\cite{swift2001trust} and the structure based on awards and sanctions are part of the accountability process. To monitor social transformations and develop a dynamic code of ethics that reflects the principles and values of society, it is necessary to call for the participation of stakeholders. However, the accountability process is often too expensive for early-stage Fintech, so it is necessary to analyze alternatives to implementing the idea of a Social License for Fintech.

Based on the assumption that artificial intelligence must have the following qualities: Ability, Benevolence, and Integrity, there are authors, like Aitken and co-authors~\cite{aitken2020establishing} (p.~03)  who argue that Fintech should seek a Social License (SL) to create engagement among your users and so that they can participate in the data processing.

The idea of Social License comes from the 90s~\cite{moffat2016sociallicense}, and its application occurred, mainly concerning environmental protection and responsibility for preserving forests. Furthermore, there are some examples of Social License use in the health industry~\cite{aitken2020establishing} (p.~03) that created a trustworthy relation between science and society.

Although the importance of the engagement between the community and Fintech, there are some challenges ~\cite{aitken2020establishing} (p.~11)), such as avoiding power asymmetry, not directing society's response, transparency about the risks that exist about the use of individuals' personal data, etc. Thus, the authors present the benefits of a Social License; however, they do not indicate how to apply it.

Consequently, the article by Aitken~\cite{aitken2020establishing} is relevant, as it provides an alternative for the population to interact with the Fintechs for the dynamic development of principles and values to be followed to build a reliable relationship. On the other hand, it doesn’t show how to implement this proposal to elaborate a Social License in the field of artificial intelligence employed by Fintech.

The idea of Social License is embedded in the concept of Corporate Social Responsibility (CSR), which consists of ``a voluntary approach that a business enterprise takes to meet or exceed stakeholder expectations by integrating social, ethical, and environmental concerns with the usual measures of revenue, profit, and legal obligation.''~\cite{BNETdictionary}. As global companies are becoming less and less interested in the voluntary adoption of CSR~\cite{wilburn2011achieving}, there is a constant political incentive for companies to recognize the need for ethical globalization and the adoption of global responsibility for their activities, such as: World Economic Forums (stated politicians that corporate leader must focus on sustainability), the Global Reporting Initiative, Mary Robinson's Realizing Rights, The Ethical Globalization Initiative, etc.

Unlike the legal license, the social license is provided by stakeholders in a continuous process of reformulation. The legal license is carried out by a public authority at a given time, and it is often expensive to obtain it and the legal analysis does not follow the market transformations~\cite{boutilier2017managing} (p.500). In addition, the sense of stakeholder can be obtained through the concept developed by the founder of stakeholder theory, Freeman~\cite{freeman2010strategic} classifies stakeholder into two groups: the group that can affect the company's activities and the group that will be affected by the company's activities. Thus, for the application of stakeholder theory in the data-driven industry, individuals and stakeholder groups should be divided into two categories:
\begin{itemize}

\item Stakeholders that affect the company's activities: traditional banks, financial institutions, government, regulatory agencies, consumers, etc.; 

\item Stakeholders who are affected by the company's activities: family members of consumers, individuals who are denied credit, holders of personal data used for ML training, relatives of those who have denied credit, startups that depend on credit, etc.

\end{itemize}

Stakeholders have a major impact on the company's activities as they form groups and organizations with connections to political actors~\cite{henry2011ideology}. Due to these coalitions, it is possible to obtain political support for the development of the company's project~\cite{bjork2013bringing}; however, it is possible to occur government pressures that affect legal licenses~\cite{boutilier2017managing}, for example. Consequently, the idea of the social license is misconstrued, as it is no longer based on the participation of all stakeholder to build a relationship of trust with companies.

An alternative to avoid the emergence of barriers to social license throughout its life cycle is the application of the Stakeholder Impact Index (SII)~\cite{olander2007stakeholder} designed for application in the construction industry, but which can be used in the data-driven industry with some adaptations that will not be presented in this article. According to Boutilier, the SII:
\begin{quote}

“It helps identify and prioritize stakeholders on the basis of their likely impact on the progress of a project. It arithmetically combines subjective estimates of several factors including the legitimacy of the groups, their power, their level of vested interest and their probability of having an impact. One factor called “position” (i.e. position on the issue of the project proceeding or being halted) is very similar to the social license. Olander showed that the approach can be useful in guiding relations with stakeholders.”

\end{quote}

Given the characteristics of the stakeholder theory and the social license, we can observe that the five ethical risks pointed out by FLORIDI~\cite{floridi2019translating} could be faced through the adoption of the idea of Social License in the case of fintech, because it is possible to elaborate an Ethics Guideline with the participation of stakeholders to avoid ethics shopping, being a great advantage, since the ethical principles and values will be constantly updated. In addition, ethics blue washing and dumping can be resolved through the transparency of fintech in obtaining the social license and by the education of stakeholder who will understand the risks and benefits arising from fintech's activities. In addition, the certification system can be created with the participation of stakeholders who could contribute to the elaboration of the evaluation criteria for obtaining certification. Consequently, there will be greater recognition and trust in the certification system and Fintech that obtain the certificate could strengthen their brand. Finally, ethics shirking will be identified mainly through the idea of accountability and transparency of Fintech's activities to its stakeholders. 

In a statement, we note that the Social License may be an alternative to address the risks of ethical malpractices by Fintech. In addition, it allows the construction of an environment of collaboration and trust between companies and stakeholders. Consequently, everyone will benefit, as Fintech will be able to scale by building a brand that is reliable and meets global ethical requirements; and stakeholders will be able to actively participate in the elaboration of ethical principles and values to be followed by companies and could have a better control of the use of their personal data.

\section{Methods and Results}

This is an exploratory study aimed at the comprehension of consumers' lack of confidence regarding Fintech's ethical concerns, especially regarding the use of Artificial Intelligence. There are not many papers considering the application of stakeholder theory and social license within Fintech. Thus, we sought to identify the variables related to this theme and verify the extent to which it is possible to use the idea of Social License to face the risks arising from the high elaboration of recommendations and Ethics Guidelines.

The comparative method was used to analyze the Social License application in the mining and construction industry and to understand the extent to which the idea can be applied in Fintech. The deductive method was used to point out the consequences resulting from the use of the theory of stakeholders and Social License's idea.  Finally, a primary research approach to data-gathering was used because there is no satisfactory existing data about this theme.

The research result was that the model of the association between Fintech is the most appropriate for collaboration between companies and for building an environment of trust between companies and stakeholders. Also, Social License is interesting for constructing the idea of a brand of trust and the participation of stakeholders in elaborating the Code of Ethics. Thus, transparency and accountability are essential in this relationship between companies and stakeholders to be possible the employment of a Social License within Fintech. 

In addition, Fintech's participation in an association allows the costs arising from the Social License employment to be distributed. Moreover, the risk of ethics shirking is dissolved because the distribution of social responsibility will be balanced within the association, and it will be possible to trace malpractices. 

Finally, the association model will allow a more effective dialogue between Fintech and governments, and it could overcome problems such as legal licenses for early-stage Fintech or the legislation change so that ML training avoids the perpetuation of discrimination in society as we verified in the study of Kelley and Ovchinnivov~\cite{kelley2020anti}.

\section{Conclusion}

The approach suggests the construction of an association formed by Fintech for stakeholder participation in elaborating, implementing, and correction of the ethical principles applicable in all fintech activities. The Social License seems to apply to the data-driven industry. It allows creating an environment of trust and dialogue, much necessary for realizing the ethical principles applicable to Artificial Intelligence. 

The risks arising from the excess of Ethics Guidelines can be faced by using the Social License and Fintech emerge as companies that need more significant participation of stakeholders to face the problem of consumer distrust and global-scale power. Moreover, mainly, due to the lack of economic resources of the early-stage Fintech, it is important to create an association in which responsibility is distributed so as not to hinder the start of the activities of the new Fintech and to facilitate the adoption of ethical principles and values from the beginning of its activities. 

Finally, new qualitative and quantitative studies will test a model of the association between Fintech, applying the stakeholder theory to the Social License in this promising economic sector. Preliminary, we consider that the Social License obtained under an association formed by Fintech is the best way to avoid cosmetic solutions regarding ethical principles and values by the data-driven industry.

\bibliographystyle{unsrt} 
\bibliography{references}

\begin{thebibliography}{10}

\bibitem{spector2006evolution}
Lee Spector.
\newblock Evolution of artificial intelligence.
\newblock {\em Artificial Intelligence}, 170(18):1251--1253, 2006.

\bibitem{crevier1993ai}
Daniel Crevier.
\newblock {\em {AI: the tumultuous history of the search for artificial
  intelligence}}.
\newblock Basic Books, Inc., 1993.

\bibitem{shapiroPaley}
Adam~R. Shapiro.
\newblock {William Paley's Lost `Intelligent Design'}.
\newblock {\em History and Philosophy of the Life Sciences}, pages 55--77,
  2009.

\bibitem{LIETO20181}
Antonio Lieto, Mehul Bhatt, Alessandro Oltramari, and David Vernon.
\newblock The role of cognitive architectures in general artificial
  intelligence.
\newblock {\em Cognitive Systems Research}, 48:1--3, 2018.

\bibitem{VANDERELST201856}
Dieter Vanderelst and Alan Winfield.
\newblock An architecture for ethical robots inspired by the simulation theory
  of cognition.
\newblock {\em Cognitive Systems Research}, 48:56--66, 2018.
\newblock Cognitive Architectures for Artificial Minds.

\bibitem{hagendorff2020ethics}
Thilo Hagendorff.
\newblock {The ethics of AI ethics: An evaluation of guidelines}.
\newblock {\em Minds and Machines}, 30(1):99--120, 2020.

\bibitem{jobin2019global}
Anna Jobin, Marcello Ienca, and Effy Vayena.
\newblock {The global landscape of AI ethics guidelines}.
\newblock {\em Nature Machine Intelligence}, 1(9):389--399, 2019.

\bibitem{cao2020ai}
Longbing Cao.
\newblock {AI in FinTech: A Research Agenda}.
\newblock {\em arXiv preprint arXiv:2007.12681}, 2020.

\bibitem{belanche2019artificial}
Daniel Belanche, Luis~V Casal{\'o}, and Carlos Flavi{\'a}n.
\newblock {Artificial Intelligence in FinTech: understanding robo-advisors
  adoption among customers}.
\newblock {\em Industrial Management \& Data Systems}, 2019.

\bibitem{park2016robo}
J.~Y. Park, J.~P. Ryu, and H.~J. Shin.
\newblock Robo advisors for portfolio management.
\newblock {\em Advanced Science and Technology Letters}, 141:104--108, 2016.

\bibitem{gai2018survey}
Keke Gai, Meikang Qiu, and Xiaotong Sun.
\newblock {A survey on FinTech}.
\newblock {\em Journal of Network and Computer Applications}, 103:262--273,
  2018.

\bibitem{giudici2018fintech}
Paolo Giudici.
\newblock Fintech risk management: A research challenge for artificial
  intelligence in finance.
\newblock {\em Frontiers in Artificial Intelligence}, 1:1, 2018.

\bibitem{bofondi2017thebig}
Giorgio Bofondi, Marcello;~Gobbi.
\newblock {The big promise of Fintech}.
\newblock {\em European Economy: banks, regulation, and the real sector},
  2:107–119, 2017.

\bibitem{mention2019future}
Anne-Laure Mention.
\newblock {The Future of Fintech}.
\newblock {\em Research-Technology Management}, 62(4):59--63, 2019.

\bibitem{bose2018world}
Anirban Bose, Penry Price, and Vincent Bastid.
\newblock World fintech report 2018.
\newblock {\em Capgemini and Linkedin Report}, 2018.

\bibitem{zetzsche2017regulating}
Dirk~A Zetzsche, Ross~P Buckley, Janos~N Barberis, and Douglas~W Arner.
\newblock Regulating a revolution: From regulatory sandboxes to smart
  regulation.
\newblock {\em Fordham J. Corp. \& Fin. L.}, 23:31, 2017.

\bibitem{dostov2017regulatory}
Victor Dostov, Pavel Shoust, and Ekaterina Ryabkova.
\newblock Regulatory sandboxes as a support tool for financial innovations.
\newblock {\em Journal of Digital Banking}, 2(2):179--188, 2017.

\bibitem{kelley2020anti}
Stephanie Kelley and Anton Ovchinnikov.
\newblock {Anti-discrimination Laws, AI, and Gender Bias in Non-mortgage
  Fintech Lending}.
\newblock {\em AI, and Gender Bias in Non-mortgage Fintech Lending}, October,
  26 2020.

\bibitem{vigdor2019apple}
Neil Vigdor.
\newblock {Apple Card Investigated after gender discrimination complaints}.
\newblock {\em The New York Times}, November, 10th. 2019 [Online].

\bibitem{aristotle1983ethics}
Aristotle.
\newblock {\em {Nicomachean Ethics}}.
\newblock Kegan Paul, Trech, Trubner, 1893.

\bibitem{winfield2019updated}
Alan Winfield.
\newblock {An updated round up of ethical principles of robotics and AI}.
\newblock
  \url{https://tex.stackexchange.com/questions/3587/how-can-i-use-bibtex-to-cite-a-web-page}.
\newblock Online; accessed: 10/feb/2021.

\bibitem{floridi2019translating}
Luciano Floridi.
\newblock {Translating principles into practices of digital ethics: Five risks
  of being unethical}.
\newblock {\em Philosophy \& Technology}, 32(2):185--193, 2019.

\bibitem{benkler2019don}
Yochai Benkler.
\newblock {Don't let industry write the rules for AI}.
\newblock {\em Nature}, 569(7754):161--162, 2019.

\bibitem{schroeder2019ethics}
Doris Schroeder, Kate Chatfield, Michelle Singh, Roger Chennells, and Peter
  Herissone-Kelly.
\newblock {Ethics Dumping and the Need for a Global Code of Conduct}.
\newblock In {\em Equitable Research Partnerships}, pages 1--4. Springer, 2019.

\bibitem{nasdaq2021shirking}
Nasdaq.
\newblock Shirking: Financial terms.
\newblock \url{https://www.nasdaq.com/glossary/s/shirking}.
\newblock Glosary. Online; accessed: 10/feb/2021.

\bibitem{ethicsGuidelines}
European Comission.
\newblock {Ethics guidelines for trustworthy AI}.
\newblock
  \url{https://ec.europa.eu/digital-single-market/en/news/ethics-guidelines-trustworthy-ai
  }.
\newblock Online; accessed: 10/feb/2021.

\bibitem{brown2006approaches}
Judy Brown and Michael Fraser.
\newblock Approaches and perspectives in social and environmental accounting:
  an overview of the conceptual landscape.
\newblock {\em Business Strategy and the Environment}, 15(2):103--117, 2006.

\bibitem{swift2001trust}
Tracey Swift.
\newblock Trust, reputation and corporate accountability to stakeholders.
\newblock {\em Business Ethics: A European Review}, 10(1):16--26, 2001.

\bibitem{aitken2020establishing}
Mhairi Aitken, Ehsan Toreini, Peter Carmichael, Kovila Coopamootoo, Karen
  Elliott, and Aad van Moorsel.
\newblock {Establishing a social licence for Financial Technology: Reflections
  on the role of the private sector in pursuing ethical data practices}.
\newblock {\em Big Data \& Society}, 7(1):2053951720908892, 2020.

\bibitem{moffat2016sociallicense}
Kieren Moffat, Justine Lacey, Airong Zhan, and Sina Leipold.
\newblock The social licence to operate: a critical review.
\newblock {\em Forestry: An International Journal of Forest Research},
  89(5):477--488, 2016.

\bibitem{BNETdictionary}
BNET~Business Dictionary.
\newblock {Corporate Social Responsability}.
\newblock
  \url{http://dictionary.bnet.com/definition/Corporate+Social+Responsibility.html},
  2009.

\bibitem{wilburn2011achieving}
Kathleen~M. Wilburn and Ralph Wilburn.
\newblock Achieving social license to operate using stakeholder theory.
\newblock {\em Journal of International Business Ethics}, 4(2), 2011.

\bibitem{boutilier2017managing}
Robert~G Boutilier and Michal Zdziarski.
\newblock Managing stakeholder networks for a social license to build.
\newblock {\em Construction Management and Economics}, 35(8-9):498--513, 2017.

\bibitem{freeman2010strategic}
R.~Edward Freeman.
\newblock {\em {Strategic management: A stakeholder approach}}.
\newblock Cambridge University Press, 2010.

\bibitem{henry2011ideology}
Adam~Douglas Henry.
\newblock Ideology, power, and the structure of policy networks.
\newblock {\em Policy Studies Journal}, 39(3):361--383, 2011.

\bibitem{bjork2013bringing}
Carwil Bjork-James.
\newblock {Bringing the fight over Bolivia’s TIPNIS Road to Washington, DC}.
\newblock {\em Carwil Without Borders}, 2013.

\bibitem{olander2007stakeholder}
Stefan Olander.
\newblock Stakeholder impact analysis in construction project management.
\newblock {\em Construction Management and Economics}, 25(3):277--287, 2007.

\end{thebibliography}

\end{document}